\documentclass[conference]{IEEEtran}
\IEEEoverridecommandlockouts
\usepackage{cite}
\usepackage{amsmath,amssymb,amsfonts}
\usepackage{algorithmic}
\usepackage[pdftex]{graphicx}
\usepackage{textcomp}
\usepackage{xcolor}
\usepackage{multirow}
\usepackage{multicol}
\usepackage{textcomp}
\usepackage{siunitx}
\usepackage{graphicx}
\usepackage{lipsum}
\usepackage{url}
\usepackage{tablefootnote}[2014/01/26]
\usepackage{threeparttable}
\usepackage[super]{nth}
\usepackage{natbib}
\usepackage{dblfloatfix}
\usepackage{tabto}
\usepackage{enumerate}

\begin{document}

\def\BibTeX{{\rm B\kern-.05em{\sc i\kern-.025em b}\kern-.08em
    T\kern-.1667em\lower.7ex\hbox{E}\kern-.125emX}}

\title{Temporal Mental Health Dynamics on Social Media\\
}

\author{\IEEEauthorblockN{\textbf{Tom Tabak$^{1}$}
\IEEEauthorblockA{$^{1}$\textit{School of Electronic Engineering and Computer Science} \\
\textit{Queen Mary University of London}\\
London, United Kingdom\\
\texttt{tabaktom360@gmail.com}}
\and
\IEEEauthorblockN{\textbf{Matthew Purver $^{1}$ $^{2}$}
\IEEEauthorblockA{$^{2}$\textit{Department of Knowledge Technologies} \\
\textit{Jo\v{z}ef Stefan Institute}\\
Ljubljana, Slovenia \\
\texttt{m.purver@qmul.ac.uk}}
}}}

\maketitle

\begin{abstract}
We describe a set of experiments for building a temporal mental health dynamics system. We utilise a pre-existing methodology for distant-supervision of mental health data mining from social media platforms and deploy the system during the global COVID-19 pandemic as a case study. Despite the challenging nature of the task, we produce encouraging results, both explicit to the global pandemic and implicit to a global phenomenon, Christmas Depression, supported by the literature. We propose a methodology for providing insight into temporal mental health dynamics to be utilised for strategic decision-making.
\end{abstract} 


\section{Introduction}

Mental health issues pose a significant threat to the general population. Quantifiable data sources pertaining to mental health are scarce in comparison to physical health data \citep{b8}. This scarcity contributes to the complexity of development of reliable diagnoses and effective treatment of mental health issues as is the norm in physical health \citep{b29}. The scarcity is partially due to complexity and variation in underlying causes of mental illness. Furthermore, the traditional method for gathering population-level mental health data, behavioral surveys, is costly and often delayed \citep{b12}.
\smallbreak
Whilst widespread adoption and engagement in social media platforms has provided researchers with a plentiful data source for a variety of tasks, including mental health diagnosis; it has not, yet, yielded a concrete solution to mental health diagnosis \citep{b26}. Conducting mental health diagnosis tasks on social media data presents its own set of challenges: The users' option of conveying a particular public persona posts that may not be genuine; sampling from a sub-population that is either technologically savvy, which may lend to a generational bias, or those that can afford the financial cost of the technology, which may lead to a demographic bias. However, the richness and diversity of the available data's content make it an attractive data source.  Quantifiable data from social media platforms is by nature social and crucially (in the context of our cases study) virtual. 
\smallbreak
Quantifiable social media data enables researchers to develop methodologies for distant mental health diagnosis and analyse different mental illnesses \citep{b25}. Distant detection and analysis enables researchers to monitor relationships of temporal mental health dynamics to adverse conditions such as war, economic crisis or a pandemic such as the Coronavirus (COVID-19) pandemic.
\smallbreak
COVID-19, a novel virus, proved to be fatal in many cases during the global pandemic that started in 2019. Governments reacted to the pandemic by placing measures restricting the movement of people on and within their borders in an attempt to slow the spread of the virus. The restrictions came in the form of many consecutive temporary policies that varied across countries in their execution. We focus on arguably the most disruptive measure: The National Lockdown. This required individuals, other than essential workers (e.g. healthcare professionals) to remain in their own homes. The lockdown enforcement varied across countries but the premise was that individuals were only permitted to leave their homes briefly for essential shopping (food and medicine). This policy had far reaching social and economic impacts: growing concern towards individuals' own and their families' health, economic well-being and financial uncertainty as certain industries (such as hospitality, retail and travel) suspended operations. As a result, many individuals became redundant and unemployed which constrained their financial resources as well as being confined to their homes, resulted in excess leisure time. These experiences along with the uncertainty of the measures' duration reflected a unique period where the general public would be experiencing a similar \emph{stressful} and \emph{anxious} period, which are both feelings associated with clinical depression \citep{b27,b28}.
\smallbreak
In this paper, we investigate the task of detecting whether a user is diagnosis-worthy over a given period of time and explore what might this appropriate time period be. We investigate the role of balance of classes in datsets by experimenting with a variety of training regimes. Finally, we examine the temporal mental health dynamics in relations to the respective national lockdowns and investigate how these temporal mental health dynamics varied across countries highly-disrupted by the pandemic.
Our main contributions in this paper are: \\ 1) We demonstrate an improvement in mental health detection performance with increasingly enriched sample representations. 2) We highlight the importance of the balance in classes of the training dataset whilst remaining aware of an approximated expected balance of classes in the unsupervised (test) dataset. 3) We analyse empirically proven relationships between populations' temporal mental health dynamics and respective national lockdowns that can be used for strategic decision-making purposes.

\section{Related research}
\label{Research}
\subsection{Natural Language Processing for Mental Health Detection}

Unlike physical health conditions that often show physical symptoms, mental health is often reflected by more subtle symptoms \citep{b25, b30}. This yielded a body of work that focused on linguistic analysis of lexical and semantic uses in speech, such as diagnosing a patient with depression and paranoia \citep{b1}. Furthermore, an examination of college students’ essays, found an increased use of negative emotional lexical content in the group of students that had high scores on depression scales \citep{b2}. Such findings confirmed that language can be an indicator of an individual’s psychological state \citep{b24} which lead to the development of Linguistic Enquiry and Word Count (LIWC) software \citep{b3,b4} which allows users to evaluate texts based on word counts in a variety of categories. More recent and larger scale computational linguistics have been applied in conversational counselling by utilising data from an SMS service where vulnerable users can engage in therapeutic discussion with counsellors \citep{b5}. For a more in-depth review of uses of natural language processing (NLP) techniques applied in mental health the reader is referred to \cite{b6}.

\subsection{Social Media as a Platform for Mental Health Monitoring}
The widespread engagement in social media platforms by users coupled with the availability of platforms’ data enables researchers to extract population-level health information that make it possible to track diseases, medications and symptoms \citep{b7}. The use of social media data is attractive to researchers not only due to its vast domain coverage but also due to the cheap methodologies by which data can be collected in comparison to previously available methodologies \citep{b8}. A plethora of mental health monitoring literature have utilised this cheap and efficient data mining methodologies from a variety of social media platforms such as: Reddit \citep{b9}, Facebook \citep{b10} and Twitter \citep{b11}. 
\smallbreak
Twitter user's engagement in the popular social media platform give way for the creation of social patterns that can be analysed by researchers, making this platform a widely used data source for data mining. Additionally, the customisable parameters querying available in the Application Programmable Interface (API) allows researchers to monitor specific populations and/or domains \citep{b12}.

\subsection{Mental Health Monitoring During COVID-19 Pandemic}
In the context of the COVID-19 pandemic, we found a handful of projects with similar intentions as our own, to monitor depression during the pandemic. \cite{b13} gather large scale, pandemic-related twitter data and infers depression based on emotional characteristics and sentiment analysis of tweets. \cite{b17} focus on detecting community level depression in Australia during the pandemic. They use the distant-supervision methodologies of \cite{b22} to gather a balanced dataset, they utilise the methodology of \cite{b8} to model the rates of depression and observing the relationship with the number of COVID-19 infections in the community. Our work differs from this in three main areas: 1) We investigate the implication of different sample representations to provide more context to our classifier. 2) We retain an imbalance in our development dataset. 3) We investigate European countries (France, Germany, Italy, Spain and the United Kingdom) that experienced a relatively high number of COVID-19 infections. 

\section{Diagnosis Classifier Experiments}
In this section we describe the data mining methodology used to build a distantly supervised dataset and the classifier experiments conducted on this dataset.
\subsection{Data}
\label{Data}
To conduct the proposed experiments, we firstly construct a distantly supervised development dataset for each country, to be used in training and validation of the classifier. The data mining methods follow the novel distant-supervision methodology proposed in \cite{b8} as it is relatively cheap but also well-structured for clinical experiments.
\smallbreak
We follow the widely-accepted methodology proposed by \citet{b14} where diagnosed (\texttt{Diagnosed}) and non-diagnosed (\texttt{Control}), groups are created. In this paper we will only be exploring depression as a mental health condition, accordingly we will have a single \texttt{Diagnosed} group for each country's development dataset. However, if multiple mental issues were to be explored, then the same number of different \texttt{Diagnosed} groups would be required for each country's dataset.
\smallbreak
\subsubsection{\texttt{Diagnosed}  Group}
\label{D_Group}
We gather 200 public tweets with a geolocation inside the country of interest, posted during a two-week period during 2019. As we are searching for a \emph{depression} \texttt{Diagnosed} tweets, this two-week period needs to be chosen strategically, as we want to capture users that have been diagnosed with depression rather than seasonal affect disorder (SAD), a separate albeit a condition with similar symptoms. Tweets collected via Twitter’s API\footnote{Twitter API: \url{https://developer.twitter.com/en/docs}}, were retrieved based on lexical content indicating that the user has history/is currently dealing with a clinical case, e.g. “I was diagnosed with depression”, rather than expressing depression in a colloquial context. Human annotators were then instructed to remove tweets that are perceived to have made a non-genuine statement regarding the users' own diagnosis, most of these were referring to a third party. Examples of genuine and non-genuine tweets encountered can be seen in Table~\ref{tab:table1}.  
\begin{table}[h]
\begin{center}
\caption{Example of annotation}
\resizebox{.5\textwidth}{!}{
\begin{tabular}{ll}
\hline
\textbf{Diagnosis indication} & \textbf{Example tweet}                                                                                                                               \\ \hline
\multirow{2}{*}{Genuine}      & \multirow{2}{*}{\begin{tabular}[c]{@{}l@{}}“I was diagnosed with severe depression and \\ went through the works of treatment for it.”\end{tabular}} \\
                              &                                                                                                                                                      \\ \hline
Non-genuine                   & \begin{tabular}[c]{@{}l@{}}“It's official. My guinea pig has\\ been diagnosed with depression”\end{tabular}                                          \\ \hline
\end{tabular}
}
\label{tab:table1}
\end{center} 
\end{table}

We then collect all (up to 5,000 most recent) tweets made public by the remaining users between the start of 2015 and October 2019. Further filtering includes removal of all users with less than 20 tweets during this period or those whose tweets do not meet our major language of instruction benchmark. This benchmark requires 70\% of the tweets collected to be written in the major language of instruction of the country of interest (i.e. United Kingdom is English, Italy is Italian etc.). Following this filtering process and some preprocessing on the tweet level, which includes medial capital splitting, mention white-space removal (i.e. if another user was mentioned this will be shown as a unique \emph{mention} token), the same has been done with URLs, all uppercase and non-emoticon related punctuation were removed.  
\smallbreak
\subsubsection{\texttt{Control} Group}
\label{C_Group}

We  gather 10,000 public tweets with a geolocation in the country of interest, posted during the same two-week period as the \texttt{Diagnosed} in 2019 and remove any tweets made by \texttt{Diagnosed} group users. We then follow a similar process to that of the \texttt{Diagnosed} collection methodology by collecting up to 5,000 most recent tweets for each user from the period mentioned above. 

\begin{table}[h]
\begin{center}
\caption{Composition of Development Datasets}
\resizebox{.5\textwidth}{!}{%
\begin{tabular}{cccc}
\hline
\textbf{Country}        & \textbf{Group} & \textbf{No. Users} & \textbf{No. Tweets} \\ \hline
\multirow{2}{*}{France} & \texttt{Diagnosed}      & $57$                 & $190,447$             \\  
                        & \texttt{Control}        & $1,041$               & $2,861,580$           \\ \hline
\multirow{2}{*}{Germany}  & \texttt{Diagnosed}      & $53$                 & $160,864$             \\  
                        & \texttt{Control}        & $1,138$               & $2,802,959$           \\ \hline
\multirow{2}{*}{Italy}  & \texttt{Diagnosed}      & $38$                 & $132,743$             \\  
                        & \texttt{Control}        & $1,051$               & $2,514,483$           \\ \hline
\multirow{2}{*}{Spain}  & \texttt{Diagnosed}     & $53$                 & $107,833$             \\  
                        & \texttt{Control}        & $1,013$               & $2,564,966$           \\ \hline
\multirow{2}{*}{U.K.}     & \texttt{Diagnosed}      & $98$                 & $289,624$             \\  
                        & \texttt{Control}        & $1,365$               & $3,319,201$           \\ \hline
\end{tabular}
}
\label{tab:table2}
\end{center}
\end{table}

As can be seen in \emph{Table~\ref{tab:table2}}, we construct imbalanced datasets. World Health Organisation (WHO) claim 264 million people suffer from depression worldwide\footnote{World Health Organization, “Depression,” 2020, [Online]. Available: \url{https://www.who.int/news-room/fact-sheets/detail/depression}[Accessed: 26 July 2020]}. Whilst, at the time of writing, the global population is approximately 7.8 billion\footnote{Worldometer. 2020. Worldometer - Real Time World Statistics. [online] Available at: \url{https://www.worldometers.info} [Accessed 19 August 2020].}. This would suggest that 1 in 30 individuals suffer from depression. However, these figures are approximations. Therefore, the extent to which our datasets are imbalanced is not an attempt to create datasets that are representative of the expected balance of classes, as these are unverifiable. Nevertheless, our datasets present ratios of \texttt{Control}:\texttt{Diagnosed} samples between $23.78$:$1$ and $11.46$:$1$, which came about from the data mining methods previously described. We accept these ratios to retain imbalanced datasets in a similar order of magnitude as the expected balance whilst achieving reasonable classifier performance.  We inherit the caveats to the distant-supervision approach of \cite{b8}: 
\begin{enumerate}[(a)]
    \item  When sampling a population we always run the risk of only capturing a subpopulation of the \texttt{Control} or \texttt{Diagnosed} that is not fully representative of the population, especially considering that \texttt{Diagnosed} samples are identified based on the fact that they publicly speak out about what is a deeply personal subject – this attribute may not generalise well to the entire population.
    \item We do not implement a verification of the method used to identify users in \texttt{Diagnosed} but rather rely on the social stigma around mental illness whereas it could be regarded as unusual for a user to tweet about a diagnosis of a mental health illness that is fictitious.
    \item \texttt{Control} is likely contaminated with users that are diagnosed with a variety of conditions, perhaps mental health related, whether they explicitly mention this or not. We have made no attempt to remove such users.
    \item Depression is often comorbid with other mental health issues \citep{b15}. As such, it is plausible that the users forming \texttt{Diagnosed} are suffering from other mental health conditions. This could suggest that the classifier will be trained to pick up these hidden meaning representations of other mental health issues and classify them as depression. We have made no attempt to further investigate nor remove such users from \texttt{Diagnosed} as having a complex \texttt{Diagnosed} group is a realistic representation of the task. 
\end{enumerate}

\subsection{Methodology}
\label{Meth}
In this section we describe the experiments conducted in training our classifier to diagnose depression. The trained classifier is deployed in \emph{Section~\ref{sec:monitor_analyse}} for classifying samples from an unsupervised experiment dataset which is then used in analysing temporal mental health dynamics.
\smallbreak
\subsubsection{Sample Representation}
\label{Sample_Rep}
We investigate the most appropriate sample representation of our distantly supervised dataset. We are posed with these considerations:
\begin{enumerate}[(a)]
\item Symptoms' temporal dependencies: as the tweets gathered come from a variety of days, weeks, months and even years, symptoms may only be present in specific time-dependant samples. However, when represented by overwhelming tweet-enriched samples the classifier performance is traded-off with retaining the symptoms' temporal dependencies. 
\item As our final task will be to monitor and analyse the temporal mental health dynamics, we are interested in modelling the rate of depression as fine-grained as possible. 
\end{enumerate}

Therefore, the ability to accurately identify \texttt{Diagnosed} samples and correctly discriminate between \texttt{Control} and \texttt{Diagnosed} with the least tweet-enriched samples will be vital in modelling a fine-grained rate of depression in the deployment stage of the final task where conclusions could be drawn in the context of the national lockdowns. The sample representations we examined: 
\begin{itemize}
    \item $Individual$ – each sample constitutes of a single tweet.
    \item $User$ $day$ -  each sample constitutes of all tweets by a unique user made public during a given day.
    \item $User$ $week$ – each sample constitutes of all tweets by a unique user made public during a given week.
    \item $All$ $user$ - each sample constitutes of all tweets made public by a unique user.
\end{itemize} 
We examine the performance of a benchmark, Support Vector Machine (SVM) with a linear kernel function \citep{b16}, on the different sample representations datasets where the benchmark classifier inputs are sparse many-hot encoding representations of the samples' lexical content.
\smallbreak
As we are working with imbalanced datasets we need to think about the metrics we use to assess the classifiers' performance. The accuracy metric is insufficient for imbalanced datasets and is best illustrated with an example. If we have a dataset with 24:1 ratio split between the samples of each class, the classifier could achieve $96\%$ accuracy by classifying every sample as the majority class. The classifier is clearly not discriminating between the distributions of the two classes but yet achieving high performance. As such, we will be assessing the performance of the classifier on the individual classes' Precision (P), Recall (R) and F1 score measures as well as the Macro F1 score for this and the remainder of the experiments in this paper. The Precision measure will tell us: of all the samples the classifier labelled as a particular class, what fraction are correct. The Recall measure will tell us: of all the samples that actually belong to that particular class, how many did the classifier correctly identify. Whilst the F1 score is a harmonic mean between the two and the Macro F1 score takes the F1 scores of all classes and calculated a non-weighted mean between them. By having a more class-specific breakdown of the classifiers' performance we can better understand the strengths and limitations of our classifiers and hence make a more informed decision when choosing the highest performing classifier.

\begin{table}[h]
\begin{center}
\caption{SVM Performance on Varying Sample Representations of U.K. Development Datasets}
\label{tab:table4}
\resizebox{.5\textwidth}{!}{
\begin{tabular}{|l|ccc|ccc|c|}
\hline
\multirow{2}{*}{\begin{tabular}[c]{@{}l@{}}\textbf{Sample}\\ \textbf{Representation}\end{tabular}} & \multicolumn{3}{c|}{\textbf{\texttt{Control}}}                                              & \multicolumn{3}{c|}{\textbf{\texttt{Diagnosed}}}                                            & \multicolumn{1}{c|}{\textbf{Macro}} \\ 
                                                                       & \multicolumn{1}{c}{P} & \multicolumn{1}{c}{R} & \multicolumn{1}{c|}{F1} & \multicolumn{1}{c}{P} & \multicolumn{1}{c}{R} & \multicolumn{1}{c|}{F1} & \multicolumn{1}{c|}{F1}    \\ \hline
\emph{Individual*}                                                             & 0.92                   & 0.99                   & 0.95                    & 0.27                   & 0.05                   & 0.08                    & 0.52                       \\ \hline
\begin{tabular}[c]{@{}l@{}}\emph{User}\\ \emph{day}\end{tabular}                     & 0.74                   & 0.96                   & 0.84                    & 0.36                   & 0.06                   & 0.1                     & 0.52                       \\ \hline
\begin{tabular}[c]{@{}l@{}}\emph{User}\\ \emph{week}\end{tabular}                    & 0.92                   & 0.97                   & 0.94                    & 0.26                   & 0.11                   & 0.15                    & 0.55                       \\ \hline
\emph{All user}                                                               & 0.94                   & 0.99                   & 0.96                    & 0.5                    & 0.14                   & 0.22                    & 0.59                       \\ \hline
\end{tabular}
}
\end{center}
\end{table}

The results in \emph{Table~\ref{tab:table4}} suggest that our benchmark classifier improved in identifying \texttt{Diagnosed}, with increasingly tweet-enriched, samples. Interestingly however, when presented with the $User$ $day$ sample representations a sharp decreased in performance in \texttt{Control} samples causing a decrease in Macro F1 score when compared with the F1 scores of both $Individual$ and $User$ $week$ sample representations. Barring this decrease in Macro F1 score, we can say that we are able to achieve improved performance when using increasingly tweet-enriched samples. However, the end task would benefit from fine-grained modelling of the rate of depression, providing us with more detailed relationships between the temporal mental health dynamics and noteworthy dates. As such, our task is bias towards the two fine-grained sample representations, $Individual$ and $User$ $day$. As our benchmark classifier achieves superior performance on the $Individual$ sample representation we will adopt this representation, as denoted by the asterisk in \emph{Table~\ref{tab:table4}}.
\\

\subsubsection{Classifier Experiments on U.K. Development Dataset}
\label{Model_Dev}

Once we have chosen the sample representation that balances out or fine-grained sample requirements with the benchmark classifier performance, we must now build a classifier that best discriminates between our two classes. The higher the performance of the classifier, the more accurate the temporal mental health dynamics will in \emph{Section~\ref{sec:monitor_analyse}}. We outline the classifier architectures included in our experimentation: 

\begin{itemize}
\item $SVM$: Linear kernel SVM as used in \emph{Section~\ref{Sample_Rep}}. This classifier will serve as our benchmark.
\item $AVEPL_{EFC}$ \footnote{All uses of $X_{EFC}$ indicate a learned embedding layer as the first layer in the classifier, after the input, and 3 Fully Connected layers following the specified classifier architecture, the activations of which are Rectified Linear (ReLU) before the output layer with a sigmoid activation.}: Average pooling layer.
\item $CNN$-$MXPL_{EFC}$: Unigram level CNN with 1 $filter$ and $kernel$ $size$ of 1 and a Max-pooling layer.
\item $BILSTM_{EFC}$: Bi-directional LSTM \citep{b18}.
\item $CNN$-$BILSTM_{EFC}$: Unigram level CNN with 1 $filter$ and $kernel$ $size$ of 1 and a Bi-directional LSTM. 
\item $CNN$-$ATT$: Unigram level CNN with 1 $filter$ and $kernel$ $size$ of 1, Attention layer \citep{b19}  and an Average pooling layer.
\item $BILSTM$-$SELFA$: Bi-directional LSTM and a Self-attention layer.
\item $BERT$: Pretrained $BERT_{Base}$\footnote{Exact pretrained $BERT_{Base}$ version implementation available here:  \url{https://tfhub.dev/google/bert_uncased_L-12_H-768_A-12/1}} fine-tuned on our dataset \citep{b20}. 
\end{itemize} 
\smallbreak
We set hyper-parameters where an Adam optimiser \citep{b21} is used with a $learning$ $rate$ of $0.01$, $batch$ $size$ of $1,000$. All classifiers were trained for a single epoch with a dataset training:validation split of $4$:$1$ and weighting the samples of \texttt{Diagnosed} as 5 times more valuable than those of \texttt{Control}. Training was done on a single Tesla P100-PCIE with 16GB of RAM available through Google's Colaboratory\footnote{Google Colaboratory: \url{https://colab.research.google.com}}.

\begin{table}[h]
\caption{Classifiers Performance on U.K. Development Dataset}
\label{UK_dev_res}
\resizebox{.5\textwidth}{!}{%
\begin{tabular}{|l|ccc|ccc|c|}
\hline
\multirow{2}{*}{\textbf{Classifier}} & \multicolumn{3}{c|}{\textbf{\texttt{Control}}}                                              & \multicolumn{3}{c|}{\textbf{\texttt{Diagnosed}}}                                            & \multicolumn{1}{c|}{\textbf{Macro}} \\ 
                            & \multicolumn{1}{c}{P} & \multicolumn{1}{c}{R} & \multicolumn{1}{c|}{F1} & \multicolumn{1}{c}{P} & \multicolumn{1}{c}{R} & \multicolumn{1}{c|}{F1} & \multicolumn{1}{c|}{F1}    \\ \hline
$SVM$                         & 0.92                   & 0.9                    & 0.91                    & 0.27                   & 0.05                   & 0.08                    & 0.52                       \\ \hline
$AVEPL$                       & 0.94                   & 0.95                   & 0.94                    & 0.33                   & 0.26                   & 0.29                    & 0.62                       \\ 
$CNN$-$MXPL$                    & 0.94                   & 0.95                   & 0.94                    & 0.31                   & 0.25                   & 0.28                    & 0.61                       \\ 
$BILSTM$                      & 0.94                   & 0.95                   & 0.94                    & 0.3                    & 0.33                   & 0.31                    & 0.63                       \\ 
$CNN$-$BILSTM$                  & 0.93                   & 0.97                   & 0.95                    & 0.3                    & 0.15                   & 0.2                     & 0.57                       \\ 
$CNN$-$ATT$                     & 0.94                   & 0.94                   & 0.94                    & 0.31                   & 0.3                    & 0.3                     & 0.62                       \\ 
$BILSTM$-$SELFA$*               & 0.94                   & 0.94                   & 0.94                    & 0.32                   & 0.31                   & 0.31                    & 0.63                       \\ \hline
$BERT$                        & 0.94                   & 0.66                   & 0.78                    & 0.12                   & 0.52                   & 0.2                     & 0.47                       \\ \hline
\end{tabular}
}
\end{table}

\smallbreak
\emph{Table~\ref{UK_dev_res}} shows that all classifiers achieve significantly higher performance on \texttt{Control} than \texttt{Diagnosed}. As we are trying to correctly detect \texttt{Diagnosed} samples and discriminate between the two classes, we prioritise the \emph{\texttt{Diagnosed} Precision} and \emph{Macro F1} score metrics. Based on these 2 chosen metrics to guide our classifier selection process 3 candidates emerge: $AVEPL$, $BILSTM$ and $BILSTM$-$SELFA$ achieving \{\emph{\texttt{Diagnosed} Precision}, \emph{Macro F1}\} scores of: \{0.33, 0.62\}; \{0.3, 0.63\} and \{0.32, 0.63\} respectively. Whilst the performance of these classifiers is similar the performance of $BILSTM$-$SELFA$ is the highest performance combination of the desired metrics (indicated by the asterisk) and as such we will be adopting this classifier in further experiments.
\\
\subsubsection{Dataset Balance Experiment}
In this section we investigate the distribution of our datasets in training and validation of our classifier. By conducting this experiment we intend to gather an in-depth understanding of our task from a linguistic standpoint. We train and validate the classifier on datasets with varying balances to investigate the role of our imbalanced dataset in the depression diagnosis task. This experiment analyses the performance of the $BILSTM$-$SELFA$ classifier on a number of different training regimes:
    \begin{itemize}
        \item $Balanced$: a dataset containing all \texttt{Diagnosed} samples and downsampling from \texttt{Control}. 
        \item $Imbalanced$:  a dataset of the development dataset's distribution (See  \emph{Table~\ref{tab:table2}}).
    \end{itemize}

Furthermore, we explore the effects of sample weighting of the classes by weighting \texttt{Diagnosed} samples as 5 times more valuable than \texttt{Control} samples as mentioned in the previous experiment. The performance of the $BILSTM$-$SELFA$ classifier on the different training regimes can be seen in \emph{Table~\ref{tab:balance}}.

\begin{table}[h]
\caption{$BILSTM$-$SELFA$ Performance on Varying Training Regimes}
\label{tab:balance}
\resizebox{.5\textwidth}{!}{%
\begin{tabular}{|l|l|l|ccc|ccc|c|}
\hline
\multirow{2}{*}{\textbf{Training}} & \multirow{2}{*}{\textbf{Validation}} & \multirow{2}{*}{\begin{tabular}[c]{@{}l@{}}\textbf{Sample} \\ \textbf{Weighting}\end{tabular}} & \multicolumn{3}{c|}{\textbf{\texttt{Control}}} & \multicolumn{3}{c|}{\textbf{\texttt{Diagnosed}}} & \textbf{Macro} \\  
                          &                             &                                                                              & P        & R       & F1      & P        & R        & F1       & F1    \\ \hline
$Balanced$                  & $Balanced$                    & None                                                                         & 0.69     & 0.69    & 0.69    & 0.69     & 0.69     & 0.69     & 0.69  \\ \hline
$Imbalanced$                & $Imbalanced$                  & None                                                                         & 0.93     & 1       & 0.96    & 0.72     & 0.11     & 0.19     & 0.58  \\
$Imbalanced$                & $Imbalanced$                  & Weighted                                                                     & 0.94     & 0.94    & 0.94    & 0.32     & 0.31     & 0.31     & 0.63  \\ \hline
$Balanced$                  & $Imbalanced$                  & None                                                                         & 0.95     & 0.63    & 0.76    & 0.14     & 0.66     & 0.23     & 0.49  \\ 
$Balanced$                  & $Imbalanced$                  & Weighted                                                                     & 0.99     & 0.13    & 0.23    & 0.09     & 0.98     & 0.16     & 0.2   \\ \hline
$Imbalanced$                & $Balanced$                    & None                                                                         & 0.53     & 1       & 0.69    & 0.98     & 0.11     & 0.2      & 0.45  \\ 
$Imbalanced$                & $Balanced$                    & Weighted                                                                     & 0.58     & 0.96    & 0.72    & 0.88     & 0.29     & 0.44     & 0.58  \\ \hline
\end{tabular}
}
\end{table}

The $Balanced$-$Balanced$ ($Training$-$Validation$) training regime achieves encouraging results in terms of its Precision-Recall trade-off, for both classes, as well as the Macro F1 score. This shows that the problem is reasonably linguistically achievable, when the imbalance challenge is removed from the equation. The $Imbalanced$-$Imbalanced$ training regime shows that adjusting the sample weighting is a successful measure we can implement to adjust the Precision-Recall trade-off in our class of interest (\texttt{Diagnosed}). Our classifier performs significantly worse in the $Balanced$-$Imbalanced$ regime when compared to the performance on the $Imbalanced$-$Imbalanced$ regime, this performance is reduced by the introduction of sample weighting. This means that when training on a $Balanced$ dataset our classifier is less robust to an $Imbalanced$ dataset at validation. Finally, whilst our classifier experiences a significant improvement in performance on the $Imbalanced$-$Balanced$ training regime when sample weighting is introduced due to our final depression diagnosis task in which we expect an $Imbalanced$ unsupervised dataset (discussed in \emph{Section~\ref{C_Group}}) the training regimes implementing $Balanced$ validation datasets are not suitable approximations of our classifier's depression diagnosis performance. Therefore, we conclude that $Imbalanced$ training, with suitable sample weighting, yields more desirable and robust depression diagnosis performance as it's able to see a broader range of data examples in training (i.e. no sub-sampling).

\subsection{Results}
\label{Results}
We train separate $BILSTM$-$SELFA$ classifiers for each of the countries' imbalanced development datasets following the $Individual$ sample representation. The test performance of these classifiers can be seen in Table~\ref{tab:Cntry_data_perf}.

\begin{table}[h]
\caption{$BILSTM$-$SELFA$ Classifier Performance on Countries' Development Datasets}
\label{tab:Cntry_data_perf}
\resizebox{.5\textwidth}{!}{
\begin{tabular}{|l|ccc|ccc|c|}
\hline
\multirow{2}{*}{\textbf{Country}} & \multicolumn{3}{c|}{\textbf{\texttt{Control}}} & \multicolumn{3}{c|}{\textbf{\texttt{Diagnosed}}} & \textbf{Macro} \\ 
                         & P        & R       & F1      & P        & R        & F1       & F1    \\ \hline
France                   & 0.96     & 0.94    & 0.95    & 0.29     & 0.37     & 0.33     & 0.64  \\ \hline
Germany                  & 0.96     & 0.96    & 0.96    & 0.3      & 0.32     & 0.31     & 0.63  \\ \hline
Italy                    & 0.97     & 0.96    & 0.97    & 0.38     & 0.4      & 0.39     & 0.68  \\ \hline
Spain                    & 0.97     & 0.97    & 0.97    & 0.35     & 0.38     & 0.36     & 0.67  \\ \hline
U.K.                     & 0.94     & 0.94    & 0.94    & 0.32     & 0.31     & 0.32     & 0.63  \\ \hline
\end{tabular}
}
\end{table}

We observe that the $BILSTM$-$SELFA$ classifier architecture achieved similar performance on the remaining countries' datasets as was acheived on U.K. dataset. This shows that the $BILSTM$-$SELFA$ architecture is able to generalise well to different languages and cultural differences after training. Hence, producing an encouraging set of results and increase our confidence in its classification ability. Whilst the $BILSTM$-$SELFA$ classifier architecture achieved the highest performance of all our classifier architectures, a combination of 0.32 \texttt{Diagnosed} Precision and 0.63 Macro F1 score leaves much to be desired. As such, we perform an error analysis and examine the significance of its results.
\\
\subsubsection{Error Analysis}
 Table~\ref{error} shows the input samples, \emph{Text}, the \emph{Prediction type} as well as the \emph{Sigmoid Output} which is the output layer of the classifier and is responsible for the final classification of the samples. The \emph{Sigmoid Output} is normalised in the range of $[0, 1] \in \mathbb{R}$, where an output of 0.5 represent the decision boundary, as such it can be interpreted as complete uncertainty by the classifier as to how the sample should be classified. A \emph{Sigmoid Output} of 1 is complete certainty that the sample should be classified as positive (\texttt{Diagnosed}) and an output of 0 is complete certainty the sample should be classified as negative (\texttt{Control}). 

\begin{table}[h]
\begin{center}
\caption{Classification Examples for Error Analysis}
\label{error}
\resizebox{.5\textwidth}{!}{
\begin{tabular}{llc}
\hline
\multicolumn{1}{c}{\textbf{\begin{tabular}[c]{@{}c@{}}Prediction\\ Type\end{tabular}}} & \multicolumn{1}{c}{\textbf{Text}}                                                                                                                                                                                                                                                            & \multicolumn{1}{c}{\textbf{\begin{tabular}[c]{@{}c@{}}Sigmoid \\ Output\end{tabular}}} \\ \hline
\begin{tabular}[c]{@{}l@{}}True \\ Positive\end{tabular}                                 & \begin{tabular}[c]{@{}l@{}}“hi davenport handmade is a small one man business i make handmade wooden bowls pens \\ jewellery boxes and other wooden items in a workshop that i built myself it started as a way \\ of overcoming depression and has taken over my life”\end{tabular} & $0.999$                                                                                   \\ \hline
\begin{tabular}[c]{@{}l@{}}False \\ Positive\end{tabular}                                & “im too depressed lol”                                                                                                                                                                                                                                                                        & $0.507$                                                                                   \\ \hline
\begin{tabular}[c]{@{}l@{}}False \\ Negative\end{tabular}                                & "i miss you too man its actually depressing me"                                                                                                                                                                                                                                               & $0.19$                                                                                    \\ \hline
\begin{tabular}[c]{@{}l@{}}True\\ Negative\end{tabular}                                  & \begin{tabular}[c]{@{}l@{}}“half term kids camps are up on wandsworth common with a dedicated kids football camp”\end{tabular}                                                                                                                                                             & $0.001$                                                                                   \\ \hline
\end{tabular}
}
\end{center}
\end{table}

We observe that the True positive example mentions having ``overcoming depression" which implies that the user has recovered from depression, as one overcomes other health issues. The \emph{Sigmoid Output} for this sample is 0.999 which is extremely high certainty by the classifier that this sample follows the distribution of \texttt{Diagnosed}. Whilst on the other end of the scale, the True negative sample discusses a topic that is completely unrelated to nor implies that the individual suffers from depression, as such it is classified as part of \texttt{Control} with a \emph{Sigmoid Output} of 0.001.
\smallbreak
However, we find the \emph{Texts} of the two samples misclassified by the classifier are rather similar. They both use words rooted from the word `depress' in rather colloquial contexts, with no indication of a past diagnosis or clinical appropriation of depression - which is rather a desirable ability for our classifier to be able to discriminate between. It is also noteworthy that the \emph{Sigmoid Outputs} of these two samples are much less polarised than the correctly classified samples, with the \emph{Sigmoid Output} of the False positive sample just about falls within the \texttt{Diagnosed} classification region. However, these two incorrectly classified samples reflect the complexity of depression diagnosis from distantly supervised tweets.
\\
\subsubsection{Significance of Results}
We investigate the significance of our classifiers' results by performing a $\chi^{2}$ significance test. Our null hypothesis, $H_{0}$, states that both sets of data, our classifiers' predictions ($\mathcal{D}_{P}$) and the distribution it is being tested against ($\mathcal{D}_{T}$), have been drawn from the same distribution ($\mathcal{D}$). 
\begin{equation}
    H_{0}: \mathcal{D}_{P} \cap \mathcal{D}_{T} \subseteq \mathcal{D}
\end{equation}
We compare the distribution of the classifiers' predictions against a random uniformly distributed set (denoted by Uniform) and against a random distributed set following the distribution of the development datasets (denoted by Weighted). All classifier results in \emph{Table~\ref{tab:Cntry_data_perf}} are statistically significant from the random baselines, according to the $\chi^{2}$ significance test - see \emph{Table~\ref{tab:Sig}} in \emph{Appendix~\ref{app:signif}}. Therefore, we can reject $H_{0}$ and conclude that the predictions of the classifier and those of the respective randomly distributed benchmarks have not been drawn from the same distribution.  

\section{Monitoring and Analysis}
We prepare the unsupervised dataset and deploy the previously trained $BILSTM$-$SELFA$ classifier to annotate this dataset. We analyse the relationships between the temporal mental health dynamics to respective national lockdown dates. 
\label{sec:monitor_analyse}
\subsection{Data}
In this section we discuss the procedure for constructing the unsupervised experiment dataset, to be used for monitoring the temporal mental health dynamics during the respective pandemic-inflicted national lockdowns.
\\
\subsubsection{Experiment Dataset}
The experiment dataset is used for the deployment of the classifier, which is trained and validated on the development set, to analyse the temporal mental health dynamics of a country. We start by gathering tweets made public by users during the first two weeks of 2020 with a geolocation within the country of interest. We then follow the same methodology for data mining as \texttt{Control} outlined in \emph{Section~\ref{Data}}, for the periods starting from 1 December 2019 until 15 May 2020, where we apply similar user-level and tweet-level preprocessing and filtering on these dataset. The composition of these experiment datasets can be seen (\emph{Table~\ref{tab:table3}} in \emph{Appendix~\ref{app:exp_dat}}) along with key dates. The key dates specified observe the official date announcing the commencement of and the announcement of first step towards easing of the national lockdowns, rather than the first official data implementing these measures as we anticipate that the announcement would provoke users to express their opinion more than the implementation of the measures. We acknowledge some caveats to the methodology with relations to the temporal mental health dynamics during the respective national lockdowns: 
\begin{enumerate}[(a)]
\item The activity-level of users whose lifestyles have been highly disrupted by the national lockdowns may be overstated during this period, due to increased leisure time.
\item The language-based filtering component may exclude certain users of the population such as stranded expatriates that use a non-majority language to communicate their thoughts. Such samples may contain a bias towards a higher rate of depression.
\end{enumerate}

\subsection{Methodology}
To monitor and analyse temporal mental health dynamics we must firstly deploy our trained $BILSTM$-$SELFA$ on the respective countries' unsupervised experiment datasets. Once we have the classifier's predictions, we must model the rate of depression by calculating the rate of depression at any given day, $R_{t}$, using the following equation:
\begin{equation}
R_{t} = \frac{\sum\limits_{i=1}^{N_{t}} \Phi(x_{i})}{N_{t}}
\end{equation}

Where $\Phi$ represents our trained classifier, $x_{i}$ is the input, $N_{t}$ is the total number of samples on day $t$. The output of the classifier, $\Phi(x_{i})$ takes the form $[0, 1] \in \mathbb{N}$. $R_{t}$ is a normalised continuous value between 0 and 1, interpreted as the proportion of tweets at $t$ that classify as \texttt{Diagnosed}: 0 meaning all samples belong to \texttt{Control} and 1 meaning all samples belong to \texttt{Diagnosed}.

\subsection{Results}
\emph{Figures~\ref{fig}} and \emph{\ref{fig1}} (see \emph{Appendix~\ref{sec:tmgd_res}}) display the temporal mental health dynamics for the countries under investigation. It is noteworthy that $R_{t}$ across the different countries is a function of the ratio of \texttt{Control}:\texttt{Diagnosed} samples in the country specific datasets on which the classifier was trained. As such, the rates across countries are not directly comparable but are rather analysed by the momentum of how $R_{t}$ in a country changed over time and how it differed from $R_{t}$ of other countries.  

\begin{figure*}[h]
  \includegraphics[width=\textwidth,height=4cm]{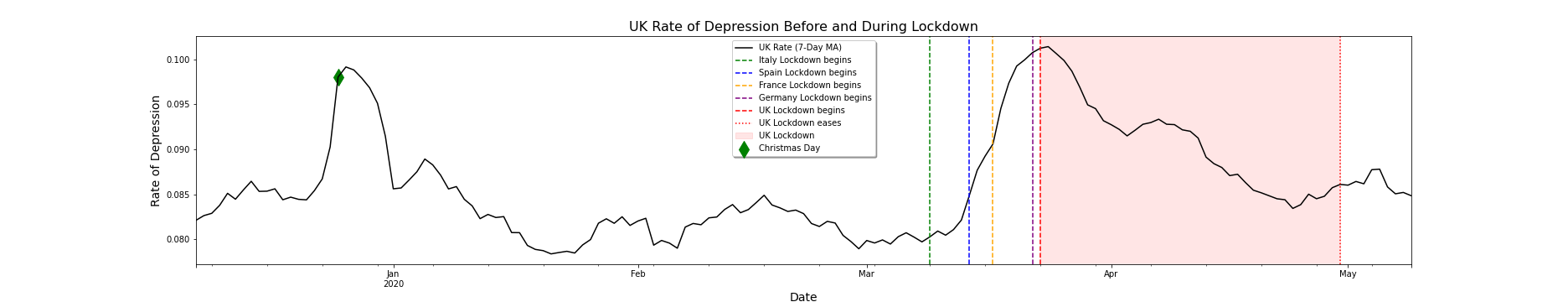}
  \caption{U.K. rate of depression before and during the National Lockdown. 
  Noise in the rate of depression has been smoothed with a 7-day moving average.}
 \label{fig}
\end{figure*}

\bigbreak
\subsection{Discussion}

Foremost, we note that we categorically cannot, nor do we, state that the rates of depression discussed in this section are \emph{caused} by the imposition of respective national lockdowns or any other measures of any type, taken by governments to combat the spread of the virus. In this section, we merely offer interpretations of the rates of depression in line with explicit relationships that we discover between the rates of depression and key events that occurred during the time-period included in this case study.
\smallbreak
Upon examination of the U.K. rate of depression ($R^{UK}$), the first distinct observation we make is not related to any pandemic-related measures but rather the sharp, non-sustained, increase of $R^{UK}$ of over 50\% on Christmas day,  before decreasing back to the status quo the next day. Upon further investigation we find that this phenomenon is well-documented \citep{b23} and seeing that our classifier was able to identify this phenomenon, without explicitly being aware of its existence, is encouraging. We continue observing $R^{UK}$ chronologically, on March \nth{9}, Italy National Lockdown begins. From this point on we observe a sharp, sustained increase in $R^{UK}$ approximately until March \nth{23}, U.K. National Lockdown begins (the last country to impose such a restriction in this study), where $R^{UK}$ somewhat plateaus. We interpret this as an increase in anxiety, a symptom of depression, amongst the U.K. population as neighbouring countries take decisive measures to slow the spread. A key theme in the build up to the U.K. national lockdown implementation was the intentional delay so that to ensure maximum utility from the policy\footnote{ITV News. 2020. Coronavirus: Boris Johnson Announces UK Government's Plan To Tackle Virus Spread, Youtube. [online] Available at: \url{https://www.youtube.com/watch?v=2U1YoKujYeY&list=PLFXSE3NhAYiZdb2qijJ7uemIB-IAYK5-y&index=893&t=0s} [Accessed 1 September 2020].}. However, a report published on the \nth{16} of March by the Imperial College COVID-19 response team\footnote{Ferguson, N., Laydon, D., Nedjati-Gilani, G., Imai, N., Ainslie, K., Baguelin, M., Bhatia, S., Boonyasiri, A., Cucunubá, Z., Cuomo-Dannenburg, G., Dighe, A., Dorigatti, I., Fu, H., Gaythorpe, K., Green, W., Hamlet, A., Hinsley, W., Okell, L., van Elseland, S., Thompson, H., Verity, R., Volz, E., Wang, H., Wang, Y., Walker, P., Walters, C., Winskill, C., Donnelly, C., Riley, Steven, R. and Ghani, A. 2020. Report 9: Impact Of Non-Pharmaceutical Interventions (Npis) To Reduce COVID-19 Mortality And Healthcare Demand. [online] Imperial.ac.uk. Available at: \url{https://www.imperial.ac.uk/media/imperial-college/medicine/sph/ide/gida-fellowships/Imperial-College-COVID19-NPI-modelling-16-03-2020.pdf} [Accessed 1 September 2020].} estimated that the the current combative approach taken up by the U.K. government would result in 250,000 deaths. The report was well-publicized by the British media and was arguably a factor in the change of combative approach by the U.K. government. This is somewhat supported by the change in $R^{UK}$ during the U.K. National Lockdown where we see a sustained \emph{decrease} for the majority of the period. Finally, we observe a slight increase in $R^{UK}$ towards the end of and in the aftermath of the National Lockdown that could perhaps be interpreted as anxiety and concern from the population at the uncertainty with which they are faced both from a social and an economic perspective. 
\smallbreak
The rates of depression of France ($R^{FR}$), Germany ($R^{DE}$), Italy ($R^{IT}$) and Spain ($R^{ES}$) behave rather differently from $R^{UK}$. Firstly, $R^{IT}$ increases sharply by over 100\% over the initial days of the Italian National Lockdown. This can be interpreted as anxiety and concern in response to the quickly imposed stringent measures by the Italian government in response to the outbreak of the virus in Italy, which at this point was largely believed to be the epicentre of the pandemic. This was coupled with economic turmoil and great concern over the capacity of hospitals and their ability to handle the high requirement for intensive care units that would ensue\footnote{CIDRAP. 2020. Doctors: COVID-19 Pushing Italian ICUs Toward Collapse. [online] Available at: \url{https://www.cidrap.umn.edu/news-perspective/2020/03/doctors-covid-19-pushing-italian-icus-toward-collapse} [Accessed 8 August 2020].}.
\smallbreak
A similar pattern emerges in $R^{FR}$, a sharp increase over the initial days of the French National Lockdown period, after-which $R^{FR}$ continues to rise throughout the lockdown period at a lower and inconsistent rate. A similar story could be tailored to $R^{ES}$. Whereas, the major increase in $R^{DE}$ occurs in the build-up month, whilst during the German National Lockdown, $R^{DE}$ increases in the initial days, albeit at a lower rate. $R^{DE}$ then plateaus and decreases - creating a turning point in $R^{DE}$ during the German National Lockdown. 
\smallbreak
Furthermore, we can observe the $R$ of the respective countries following the easing of respective lockdowns and interpret them as the countries' outlook on the easing of restrictions. From the time period that we have available it seems that the French and Spanish general populations experienced a reduction in symptoms of depression, such as anxiety, as is evidenced by the clear reduction in $R^{FR}$ and $R^{ES}$ respectively. We can therefore conclude by, tentatively, stating that the easing of restrictions were received by an improvement in the mental state of the general populations of France and Spain, the mental state of the Italian and German general populations deteriorated, whilst the general mental state of the U.K. was rather agnostic to the easing of restriction.
\smallbreak
We are hesitant to state the changes in the rates of depression had been \emph{caused} by the imposition/easing of national lockdowns. To make such a claim we would be required to undertake a more fine-grained causality study which is beyond the scope of this paper, however we note this for future work. We \emph{can} however claim to have discovered clear relationships between the drastic changes in the behaviour of rates of depression during the periods of the build-up to, during and in the aftermath of national lockdowns

\section{Conclusion}
Our set of experiments have been conducted with the aim of providing organisations with a methodology for monitoring and analysing temporal mental health dynamics using social media data. We examine sample representations and their ability to impact classifier performance. We investigate the role of including an imbalanced dataset in the classifier training regime. Our classifier provides encouraging performance on two fronts: the first being that it is able to discriminate with high certainty between clear \texttt{Diagnosed} and \texttt{Control} samples and secondly, it was able to identify the Christmas Depression phenomenon supported by the literature. Finally, we present an analysis and discussion of the rates of depression and their relationships with key events during the COVID-19 pandemic. We reiterate that through the analysis conducted in this paper, we cannot state that the measures imposed \emph{caused} the changes in rates of depression during the pandemic and leave this causality analysis for future work. Mental health monitoring methodologies such as the one proposed in this paper can be adopted by governments, to identify relationships between the general population's mental health state to imposed measures, mental health authorities, to assist in planning and targeting individual locations in which to dynamically concentrate their resources, as well corporations involved in producing or disseminating drugs, such as pharmaceuticals, to combat mental health issues for a more commercial use case.

\section*{Acknowledgments}
Purver is partially supported by the EPSRC under grant EP/S033564/1, and by the European Union’s Horizon 2020 pro- gramme under grant agreements 769661 (SAAM, Supporting Active Ageing through Multimodal coaching) and 825153 (EMBEDDIA, Cross-Lingual Embeddings for Less-Represented Languages in European News Media). The results of this publication reflect only the authors’ views and the Commission is not responsible for any use that may be made of the information it contains. We express our thanks to all of our data annotators: L. Achour, L. Del Zombo, N. Fiore, M. Hechler and R. Medivil Zamudio.

\bibliographystyle{agsm}
\bibliography{main}
\vspace{12pt}

\clearpage

\begin{appendix}

\subsection{Significance of Results}
\label{app:signif}

\begin{table}[h]
\caption{Significance in Predictions of $BILSTM$-$SELFA$ Classifier}
\resizebox{.5\textwidth}{!}{%
\begin{tabular}{llc}
\hline
\textbf{Country}        & \textbf{Comparison} & \multicolumn{1}{l}{\textbf{Significance}}  \\ \hline
\multirow{2}{*}{France} & Uniform              & $\chi^2=1,311,459$ ($p$ \textless $0.00001$)                                                  \\ 
                        & Weighted     & $\chi^2=6,726$ ($p$ \textless 0.00001)                                                     \\ \hline
\multirow{2}{*}{Germany}  & Uniform              & $\chi^2=1,504,290$ ($p$ \textless $0.00001$)                                                  \\ 
                        & Weighted     & $\chi^2=5,607$ ($p$ \textless $0.00001$)                                                     \\ \hline
\multirow{2}{*}{Italy}  & Uniform              & $\chi^2=1,485,204$ ($p$ \textless $0.00001$)                                                  \\ 
                        & Weighted     & $\chi^2=5,050$ ($p$ \textless $0.00001$)                                                     \\ \hline
\multirow{2}{*}{Spain}  & Uniform             & $\chi^2=2,122,253$ ($p$ \textless $0.00001$)                                                  \\ 
                        & Weighted     & $\chi^2=6,324$ ($p$ \textless $0.00001$)                                                     \\ \hline
\multirow{2}{*}{U.K.}     & Uniform              & $\chi^2=1,242,848$ ($p$ \textless $0.00001$)                                                  \\ 
                        & Weighted     & $\chi^2=16,591$ ($p$ \textless 0.00001)                                                     \\ \hline
\end{tabular}
}
\label{tab:Sig}
\end{table}

\subsection{Experiment Dataset Composition}
\label{app:exp_dat}

\begin{table}[h]
\begin{center}
\begin{threeparttable}
\caption{Composition of Experiment Datasets}
\begin{tabular}{lllll}
\hline
\textbf{Country} & \textbf{\begin{tabular}[c]{@{}l@{}}Restrictions \\ Begin\end{tabular}} & \textbf{\begin{tabular}[c]{@{}l@{}}Restrictions \\ Eased\end{tabular}} & \textbf{\begin{tabular}[c]{@{}l@{}}No. \\ Users\end{tabular}} & \textbf{\begin{tabular}[c]{@{}l@{}}No. \\ Tweets\end{tabular}} \\ \hline
France           & 17 March $2020^{1a}$                                                          & 11 May $2020^{2a}$                                                         & $1,351$                                                            & $945,919$                                                             \\ \hline
Germany            & 22 March $2020^{3a}$                                                           & 6 May $2020^{4a}$                                                         & $1,643$                                                            & $998,248$                                                             \\ \hline
Italy            & 9 March $2020^{5a}$                                                           & 27 April $2020^{6a}$                                                         & $1,725$                                                            & $764,089$                                                             \\ \hline
Spain            & 14 March $2020^{7a}$                                                        & 28 April $2020^{8a}$                                                         & $2,060$                                                         & $1,012,847$                                                      \\ \hline
U.K.               & 23 March $2020^{9a}$                                                           & 30 April $2020^{10a}$                                                           & $2,883$                                                         & $2,050,554$                                                      \\ \hline

\end{tabular}
\label{tab:table3}
\begin{tablenotes}
\item[1a] The Independent. 2020. France Imposes 15-Day Lockdown And Mobilises 100,000 Police To Enforce Coronavirus Restrictions. [online] Available at: \url{https://www.independent.co.uk/news/world/europe/coronavirus-france-lockdown-cases-update-covid-19-macron-a9405136.html} [Accessed 12 July 2020].
\item[2a] BBC News. 2020. France Eases Lockdown After Eight Weeks. [online] Available at: \url{https://www.bbc.co.uk/news/world-europe-52615733} [Accessed 12 July 2020].
\item[3a] BBC News. 2020. Germany Bans Groups Of More Than Two To Curb Virus. [online] Available at: \url{https://www.bbc.co.uk/news/world-europe-51999080} [Accessed 4 August 2020].
\item[4a] BBC News. 2020. Germany Says Football Can Resume And Shops Reopen. [online] Available at: \url{https://www.bbc.co.uk/news/world-europe-52557718} [Accessed 4 August 2020].
\item[5a] CNN. 2020. All Of Italy Is In Lockdown As Coronavirus Cases Rise. [online] Available at: \url{https://edition.cnn.com/2020/03/09/europe/coronavirus-italy-lockdown-intl/index.html} [Accessed 12 July 2020].
\item[6a] BBC News. 2020. Coronavirus: Italy's PM Outlines Lockdown Easing Measures. [online] Available at: \url{https://www.bbc.com/news/amp/world-europe-52435273} [Accessed 12 July 2020].
\item[7a] The Guardian. 2020. Spain Orders Nationwide Lockdown To Battle Coronavirus. [online] Available at: \url{https://www.theguardian.com/world/2020/mar/14/spain-government-set-to-order-nationwide-coronavirus-lockdown} [Accessed 12 July 2020].
\item[8a] BBC News. 2020. Spain Plans Return To 'New Normal' By End Of June. [online] Available at: \url{https://www.bbc.co.uk/news/world-europe-52459034} [Accessed 12 July 2020].
\item[9a] BBC News. 2020. Coronavirus Updates: 'You Must Stay At Home' UK Public Told - BBC News. [online] Available at: \url{https://www.bbc.co.uk/news/live/world-52000039} [Accessed 12 July 2020].
\item[10a] BBC News. 2020. UK Past The Peak Of Coronavirus, Says PM. [online] Available at: \url{https://www.bbc.co.uk/news/uk-52493500} [Accessed 12 July 2020].
\end{tablenotes}
\end{threeparttable}
\end{center}
\end{table}

\clearpage

\subsection{Temporal Mental Health Dynamics Results}
\label{sec:tmgd_res}

\begin{figure}[h]
 \centering
  \includegraphics[width=\textwidth,height=4cm]{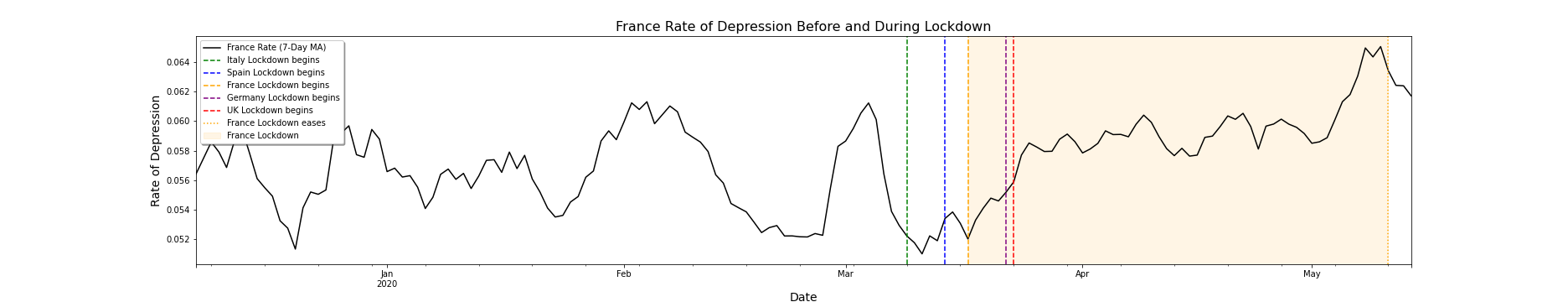}
  \includegraphics[width=\textwidth,height=4cm]{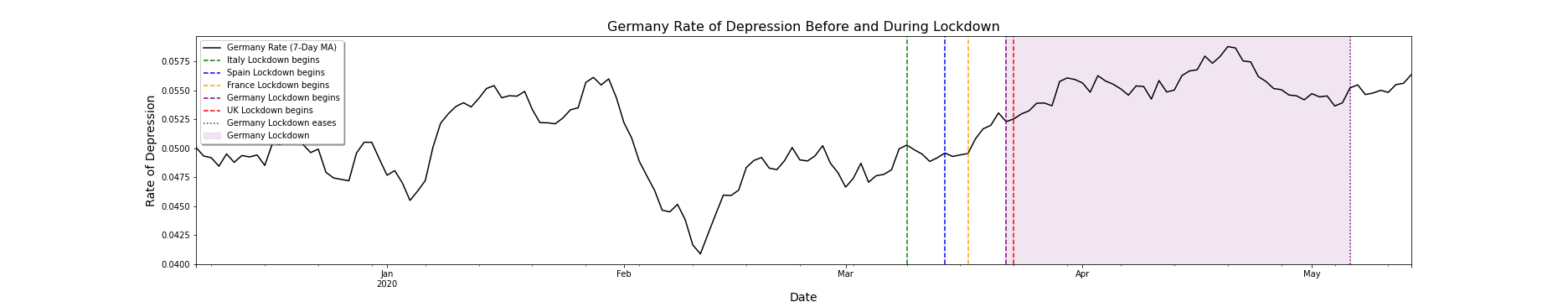}
  \includegraphics[width=\textwidth,height=4cm]{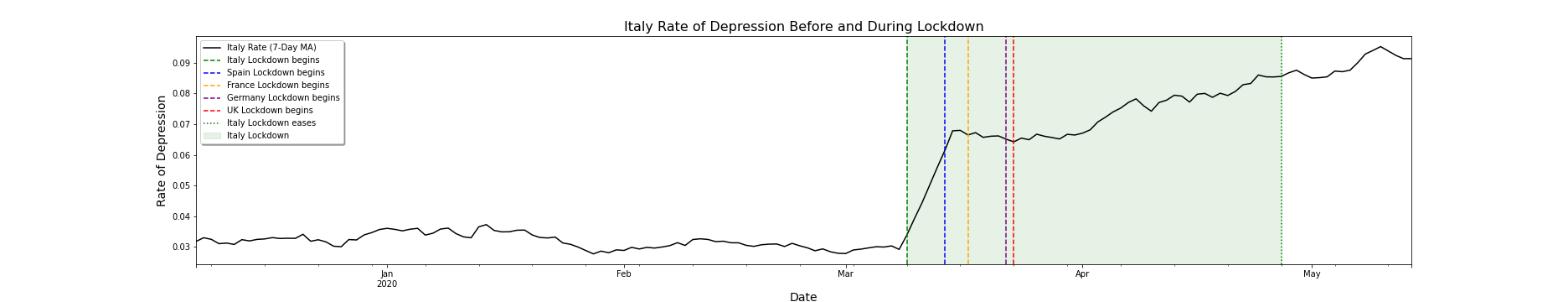}
  \includegraphics[width=\textwidth,height=4cm]{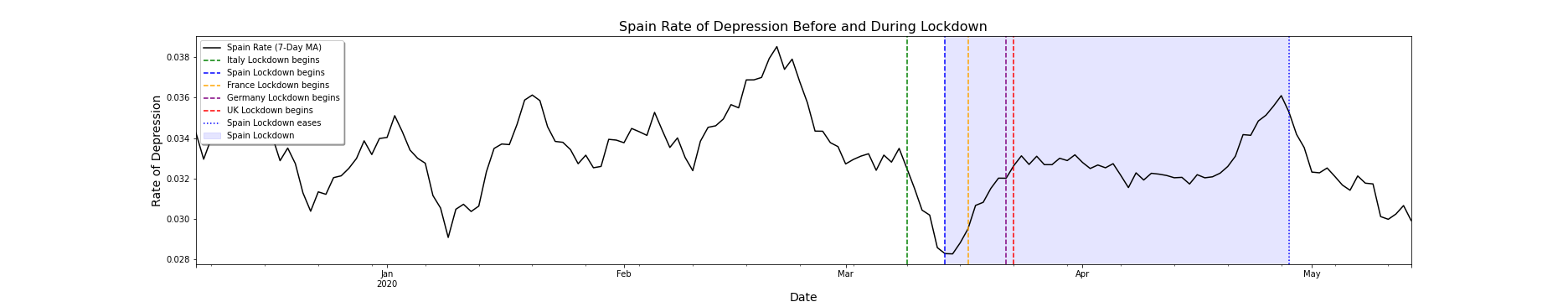}
  \caption{France, Germany, Italy and Spain rates of depression before and during respective national lockdowns. Noise in rates of depression have been smoothed with 7-day moving averages}
  \label{fig1}
\end{figure}


\end{appendix}
    
\end{document}